\documentclass{vgtc}                          

\graphicspath{{figures/}{pictures/}{images/}{./}} 

\usepackage{times}                     

\usepackage{tabu}                      
\usepackage{booktabs}                  
\usepackage{lipsum}                    
\usepackage{mwe}                       
\usepackage{color,soul}
\usepackage{mathptmx}                  

\onlineid{0}

\vgtccategory{Research}

\vgtcinsertpkg

\title{AdaVLN: Towards Visual Language Navigation in Continuous Indoor Environments with Moving Humans 
}

\author{Dillon LOH\thanks{e-mail: dillon.loh@hiverlab.com}\\ %
        \scriptsize Hiverlab Pte. Ltd.%
       \and Tomasz BEDNARZ\thanks{e-mail: tbednarz@nvidia.com}\\ %
        \scriptsize NVIDIA%
       \and Xinxing XIA\thanks{e-mail: lygxia@gmail.com}\\ %
        \scriptsize Shanghai University%
       \and Frank GUAN\thanks{e-mail: frank.guan@singaporetech.edu.sg}\\ %
     \scriptsize Singapore Institute of Technology}

\teaser{
  \centering
  {\normalsize \textbf{Project Page:} \href{https://github.com/dillonloh/AdaVLN}{github.com/dillonloh/AdaVLN}} \\
  \vspace{0.3cm}
  \includegraphics[height=0.42\linewidth,keepaspectratio]{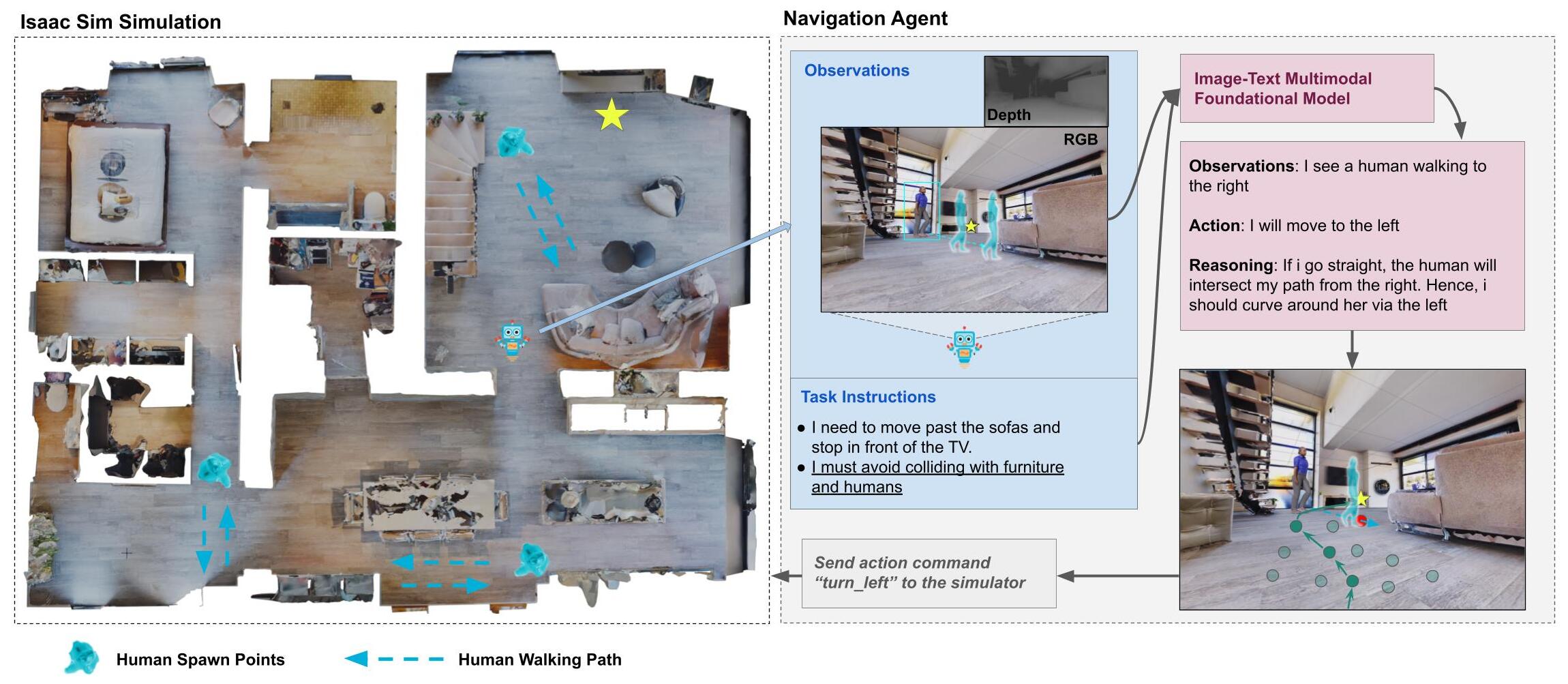}
  \caption{AdaVLN: Given natural language instructions, the robot is tasked to navigate in a continuous indoor space while avoiding collisions with moving humans. At each navigation step, the navigation agent receives an egocentric forward-facing 115-degree RGB-D observation of its environment. Our baseline agent then sends the observations and task instructions to a GPT-4o-mini model, which generates a set of observations and actions + reasoning. The actions are then sent to the simulator to execute the next navigation step.}
  \label{fig:teaser}
}

\abstract{
Visual Language Navigation is a task that challenges robots to navigate in realistic environments based on natural language instructions. While previous research has largely focused on static settings, real-world navigation must often contend with dynamic human obstacles. Hence, we propose an extension to the task, termed Adaptive Visual Language Navigation (AdaVLN), which seeks to narrow this gap. AdaVLN requires robots to navigate complex 3D indoor environments populated with dynamically moving human obstacles, adding a layer of complexity to navigation tasks that mimic the real-world. To support exploration of this task, we also present AdaVLN simulator and AdaR2R datasets. The AdaVLN simulator enables easy inclusion of fully animated human models directly into common datasets like Matterport3D. We also introduce a "freeze-time" mechanism for both the navigation task and simulator, which pauses world state updates during agent inference, enabling fair comparisons and experimental reproducibility across different hardware. We evaluate several baseline models on this task, analyze the unique challenges introduced by AdaVLN, and demonstrate its potential to bridge the sim-to-real gap in VLN research.

} 


\begin{document}


\firstsection{Introduction}

\maketitle

Visual navigation in indoor environments is a topic within the field of Embodied AI research which focuses on an agent/robot's ability to follow instructions to navigate within unknown environments towards a goal. Approaching this problem typically requires an agent/robot to: 1) understand and remember the environment its been placed in; 2) interpret natural language instructions; and 3) use information from the formers to decide on a series of actions to adequately follow the instructions given to it  \cite{Zhang_Ma_Li_Qiao_Wang_Chai_Wu_Bansal_Kordjamshidi_2024}.

While the premise of this task is straightforward, different variants have been introduced over the years, which can be broadly classified based on \textit{communication complexity} (single/multi-turn interaction), \textit{task objective} (action/goal-directed), and \textit{action space} (discrete/continuous spaces) \cite{Zhang_Ma_Li_Qiao_Wang_Chai_Wu_Bansal_Kordjamshidi_2024, Gu_Stefani_Wu_Thomason_Wang_2022, Anderson_Wu_Teney_Bruce_Johnson_Sünderhauf_Reid_Gould_Hengel_2018, Raychaudhuri_Ta_Ashton_Chang_Wang_Bucher_2024}.

Within this framework, Visual Language Navigation (VLN) is generally a single-turn, action-directed task, with discrete or continuous action spaces depending on the task variant \cite{Anderson_Wu_Teney_Bruce_Johnson_Sünderhauf_Reid_Gould_Hengel_2018, Krantz_Wijmans_Majumdar_Batra_Lee_2020}. VLN in continuous indoor environments (VLN-CE) has gained significant attention recently due to its alignment with increasingly probable real-world applications, such as home robotic assistants \cite{Krantz_Wijmans_Majumdar_Batra_Lee_2020}. However, the commonly used task datasets and simulators for existing VLN tasks are largely static and lack the dynamic complex features in real-world scenarios, such as moving obstacles and changing spaces. In real settings, humans and other entities often move within the same space that a robot is navigating, which require agents to not only follow instructions but also predict the future positions of these dynamic obstacles and adjust their routes accordingly at inference time. These are capabilities essential for success in other related navigation tasks like SOON, HANNA, VLNA, and VDN. \cite{Zhu_Liang_Zhu_Chang_Liang_2021, Wijmans_Datta_Maksymets_Das_Gkioxari_Lee_Essa_Parikh_Batra_2019, Nguyen_Dey_Brockett_Dolan_2019, Qi_Wu_Anderson_Wang_Wang_Shen_Hengel_2020, Thomason_Murray_Cakmak_Zettlemoyer_2019}.

To narrow this gap, we introduce Adaptive Visual Language Navigation (AdaVLN), a task extension of the VLN-CE problem, to incorporate moving human obstacles into the widely-used Habitat-Matterport3D environments \cite{Ramakrishnan_Gokaslan_Wijmans_Maksymets_Clegg_Turner_Undersander_Galuba_Westbury_Chang_2021}. Additionally, we propose a "freeze-time" mechanism for simulations in AdaVLN, where the otherwise constantly-running simulation is paused while the agent is processing decisions, ensuring fair comparisons across varying hardware speeds. 

\begin{figure}
    \centering
    \includegraphics[width=\columnwidth]{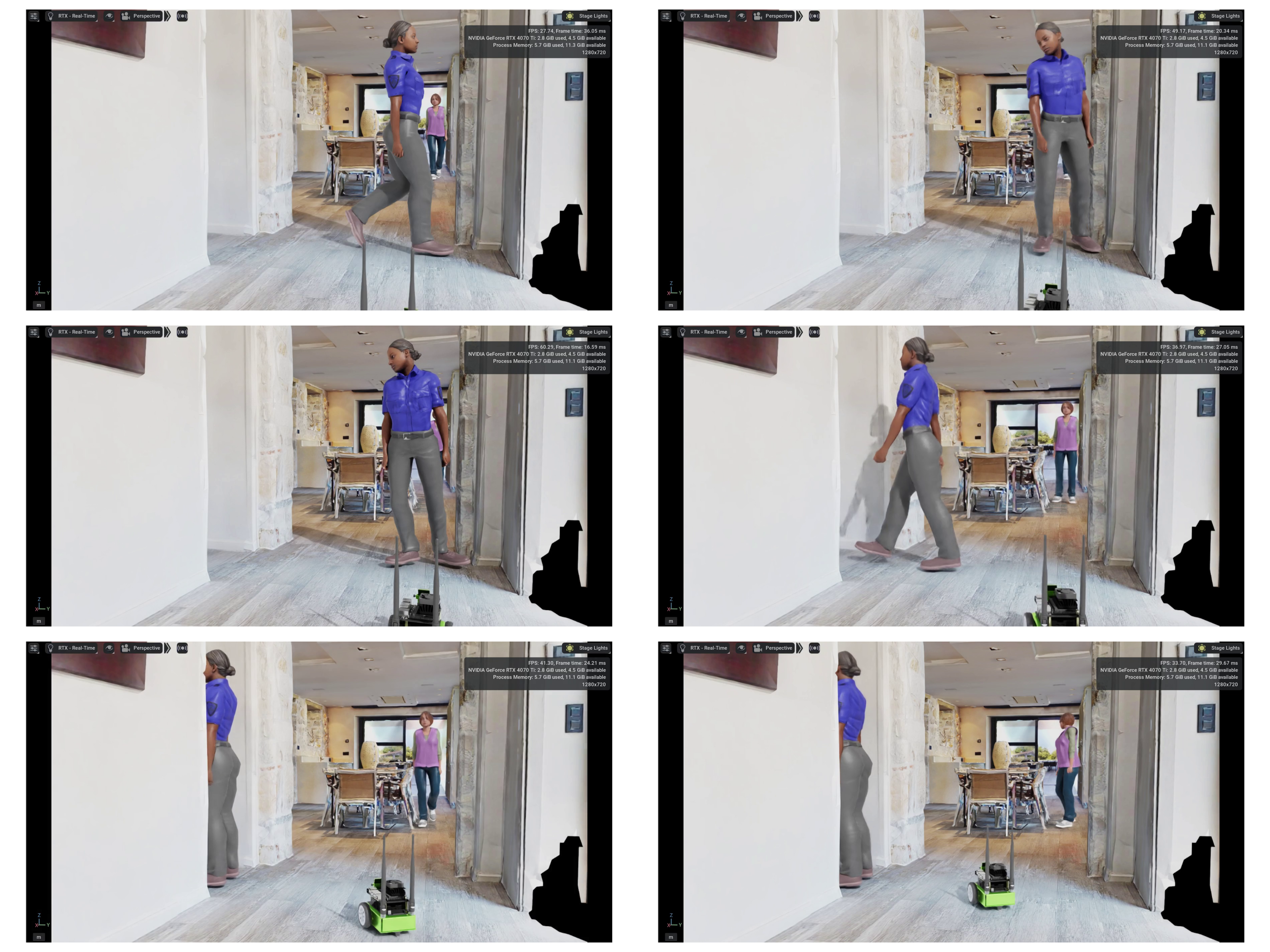}
    \caption{Jetbot navigating in a dynamic Matterport3D environment with moving human obstacles.}    
    \label{fig:sim_animated_perspective_example}
\end{figure}

\begin{figure}[t]
    \centering
    \includegraphics[width=\columnwidth]{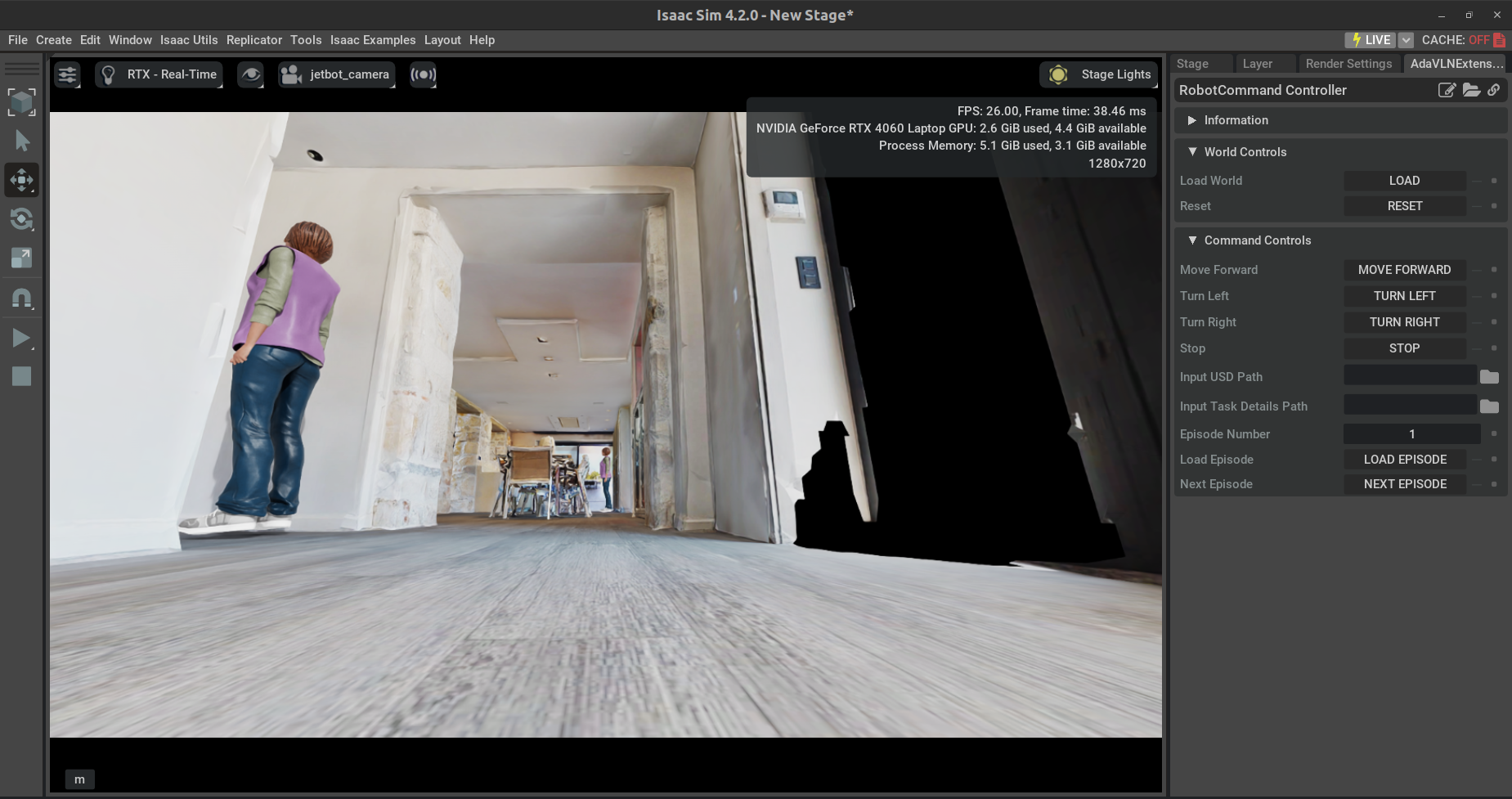}
    \caption{AdaSimulator's GUI Extension in Isaac Sim}
    \label{fig:gui}
\end{figure}

Specifically, we introduce the following two tools to enable research in this topic:
\textbf{AdaSimulator}: A simulator offering physics-based 3D environments with dynamically moving obstacles like humans and accurate mobile robot movements (refer to Figure \ref{fig:sim_animated_perspective_example} and \ref{fig:gui}). AdaSimulator is based on IsaacSim \cite{makoviychuk2021isaacgymhighperformance} and is built to be compatible with Matterport3D environments \cite{Chang_Dai_Funkhouser_Halber_Nießner_Savva_Song_Zeng_Zhang_2017} and supports easy customisation of human spawn points, and pathing logic.
\textbf{AdaR2R}: A sample variant of the R2R \cite{Anderson_Wu_Teney_Bruce_Johnson_Sünderhauf_Reid_Gould_Hengel_2018} and Matterport3D datasets that includes spawn points and trajectories for dynamic obstacles, adding another layer of realism to the navigation task.

Finally, we conduct experiments with several baseline models to evaluate our new task, analyzing the impact of these additional complexities on agent behavior and performance.

In summary, our contributions are as follows (refer to Figure \ref{fig:teaser}):
\begin{enumerate}
    \item We introduce the AdaVLN task, a variant of VLN-CE with dynamic human obstacles moving in 3D environments in order to mimic the real-life scenarios. 
    \item We introduce the AdaSimulator, an IsaacSim-based simulator that supports physics-enabled meshes and animated humans, and AdaR2R (Sample), an example dataset based on R2R-CE that enables configuration of the above-mentioned environments.
    \item We run a baseline agent based on foundational models and discuss the new difficulties agents will face when navigating in this new realistic environment. 
\end{enumerate}

\section{Related Works}

\subsection{Visual Language Navigation}

Over the years, the VLN task has evolved and produced several variants, generally aiming to narrow the gap between simulated environments and the real-world scenarios that an actual robot might encounter. The original Visual Language Navigation task and Room-to-Room (R2R) dataset were introduced by Anderson et al. \cite{Anderson_Wu_Teney_Bruce_Johnson_Sünderhauf_Reid_Gould_Hengel_2018} and requires robots to navigate in static 3D environments given a single initial instruction. At every navigation step, the robot is provided with a 360-degree panoramic RGB-D view of its surroundings, and has to choose from pre-determined neighbouring nodes to teleport to. It popularised the use of the Matterport3D scan dataset \cite{Chang_Dai_Funkhouser_Halber_Nießner_Savva_Song_Zeng_Zhang_2017} as a source of realistic 3D environments and provided the original Matterport3D Simulator. This simulator was later adapted to similar tasks in static environments, such as Scenario-Oriented Object Navigation (SOON) \cite{Zhu_Liang_Zhu_Chang_Liang_2021} and Remote Embodied Visual Referring Expressions (REVERIE) \cite{Qi_Wu_Anderson_Wang_Wang_Shen_Hengel_2020}.

Soon after, expansions to the original R2R datasets, e.g. R4R \cite{Jain_Magalhaes_Ku_Vaswani_Ie_Baldridge_2019}, RxR \cite{Ku_Anderson_Patel_Ie_Baldridge_2020}, were created to diversify and increase the difficulty of the navigation tasks. In parallel, new tasks emerged that shifted the focus to different complexities and problems within the field of visual navigation. The Embodied Question Answering (EQA) task \cite{Wijmans_Datta_Maksymets_Das_Gkioxari_Lee_Essa_Parikh_Batra_2019} and Vision-and-Language Navigation with Actions (VLNA) \cite{Nguyen_Dey_Brockett_Dolan_2019} were introduced as related tasks in which agents were challenged not only to navigate but also to answer questions or perform actions based on the visual scene.

The Habitat Sim simulator \cite{puig2023habitat3, szot2021habitat, habitat19iccv} and Habitat-Matterport3D \cite{Ramakrishnan_Gokaslan_Wijmans_Maksymets_Clegg_Turner_Undersander_Galuba_Westbury_Chang_2021} mesh datasets were later introduced, which provided a framework for conducting experiments in full physics-enabled 3D environments. Krantz el al \cite{Krantz_Wijmans_Majumdar_Batra_Lee_2020} integrated this to extend the VLN task to continuous action spaces (VLN-CE), where robots had to navigate by making 'low-level' movement decisions (turn left/right 15 degrees, move forward 0.25m, stop). RxR-Habitat competitions focusing on the VLN-CE task have since been organised in multiple years of the CVPR conference, which implement slight variants to the task parameters \cite{habitatrxrcompetition, an20221stplacesolutionsrxrhabitat}. This task further closed the sim2real gap, and has led to research into new world state modelling techniques \cite{an2024etpnavevolvingtopologicalplanning, wang2023graphbasedenvironmentrepresentation, wang2023gridmmgridmemorymap} and long-term navigation planning ideas \cite{wang2024lookaheadexplorationneuralradiance, Wang_Liang_Van_Gool_Wang_2023} specific to it. \cite{hong2022bridginggaplearningdiscrete} also showed the natural link between the discrete and continuous variants, and how they can complement each other when solving either task.

Recent work by Li et al. \cite{Li_Li_Cheng_Dong_Zhou_He_Dai_Mitamura_Hauptmann_2024} introduced the Human-Aware MP3D (HA3D) simulator for discrete action spaces, with perspective-specific animations of humans in Matterport3D environments and a corresponding Human-Aware R2R dataset, requiring robots to parse these dynamic elements as part of their navigation instructions. The paper also introduced the inclusion of collision statistics into the metrics of VLN experiments like success rate etc.

\subsection{Collision Avoidance during Navigation}

Collision avoidance in robotics is a well-researched topic, where the idea of local/offline path-planning is especially important since robots are expected to perform well in unknown environments. This typically requires the model to predict how their world state will change along a trajectory in order to preemptively avoid obstacles when planning their path. Traditional methods in robotics research for choosing safe paths used Velocity Obstacles \cite{Fiorini_Shiller_1998} to calculate potential collision paths based on both the movement of surrounding objects and the robot itself. Reciprocal Velocity Obstacles extended this to multi-agent scenarios common in real-world scenarios. Methods that extract motion information from RGB-D data was also demonstrated in \cite{Herbst_Ren_Fox_2013}. Newer prediction methods also made use of reinforcement learning techniques, models dynamic obstacles as variable-sized ellipsoids, and considers agent-human interaction dynamics to ground path predictions \cite{Truong_Ngo_2017, Pfeiffer_Shukla_Turchetta_Cadena_Krause_Siegwart_Nieto_2018, Castillo-Lopez_Sajadi-Alamdari_Sanchez-Lopez_Olivares-Mendez_Voos_2018}. These were typically combined with grid-based \cite{Elfes_2013}, graph-based \cite{Wang_Wu_Yao_Wang_2023}, or 3D-based methods \cite{Hornung_Wurm_Bennewitz_Stachniss_Burgard_2013} of modelling the world on-the-fly provided these robots with rich representations of its surroundings for better motion prediction and environment understanding.

Visual Language Navigation has since integrated many of the above ideas for future planning and obstacle prediction. Notably, DREAMWALKER \cite{Wang_Liang_Van_Gool_Wang_2023} introduced the use of mental simulations in order to predict the environment along candidate trajectories. \cite{an2024etpnavevolvingtopologicalplanning} focused on obstacle avoidance using dynamic topological planning, and later on Jeong et al. \cite{Jeong_Kang_Kim_Zhang_2024} introduced the VLN-CM agent, which leverages depth maps to predict expected occupancy maps along their candidate trajectories.

\section{Adaptive Visual Language Navigation}

The existing VLN and VLN-CE tasks are largely focused on navigation in static environments, and do not explicitly define scenarios where dynamic obstacles like moving humans are present. To provide realistic, human-populated environments, we introduce Adaptive Visual Language Navigation (AdaVLN), an extension to the VLN-CE task. 

\subsection{Task Description}
Building upon the VLN-CE task, AdaVLN sets robots in Matterport3D environments with continuous action spaces. At the start of each navigation episode (time \(t_0=0\)), the robot is initialized at position \((X_0, \theta_0)\) and is required to navigate to a goal position \(X_G\) by following a sequence of natural language instructions provided at the start. A key addition in AdaVLN is the inclusion of dynamic obstacles — in the form of humans — and an emphasis on collision avoidance. The states of these obstacles, denoted \((X'_t, \theta'_t)\), are continuously updated as they move along NavMesh paths between pre-defined waypoints in the AdaR2R dataset. Robots are required to avoid collision with both static obstacles (e.g. environment meshes) and the dynamic obstacles.

\subsection{Observations/Actions of Robots}
\begin{figure}
    \centering
    \includegraphics[width=\columnwidth]{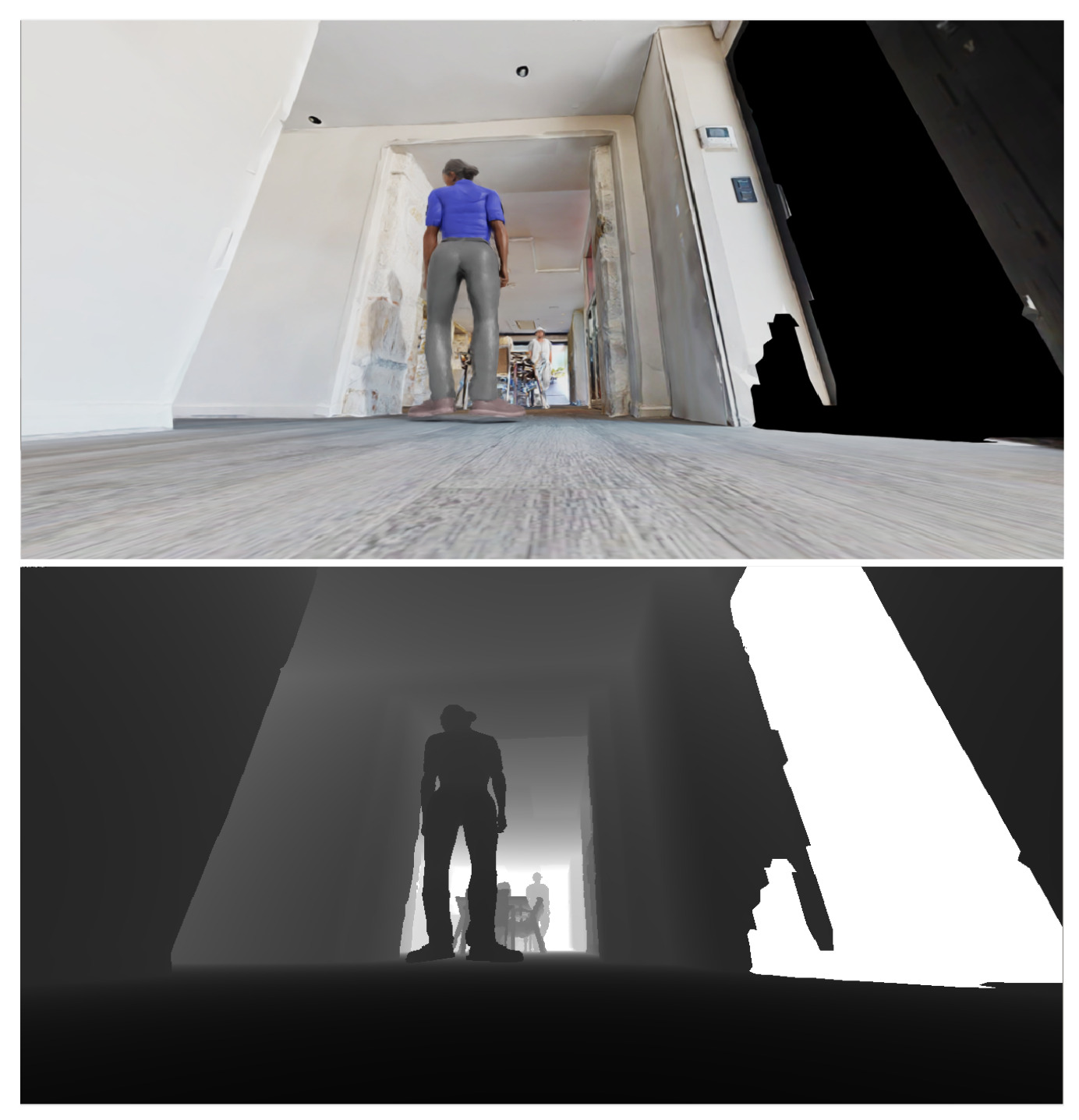}
    \caption{Top: RGB observations, Bottom: Depth observations provided to agent.  Note that the depth observations have been restricted to a range between 0 and 10 in this image for clarity.}
    \label{fig:observations}
\end{figure}

At each navigation step \(t\), the robot observes an egocentric 115-degree front-facing view of its surroundings in the form of an RGB-D image \cite{Li_Li_Cheng_Dong_Zhou_He_Dai_Mitamura_Hauptmann_2024}, as seen in Figure \ref{fig:observations}. Based on this observation and its current state, the robot can choose from one of four possible actions:

\begin{enumerate}
    \item Turn left by 15 degrees at 30 degrees/s
    \item Turn right by 15 degrees at 30 degrees/s
    \item Move forward 0.25 meters at 0.5m/s
    \item Stop
\end{enumerate}

The significance of time means that the speed at which the above actions are performed will affect the results. Our robots are configured to move with linear speeds of 0.5 m/s and rotate at 30 degrees/s. These timings were chosen to standardise the time taken by each action to 2 seconds.

The 'stop' command indicates the end of an episode, upon which the robot and simulation stops. The agent's performance is then evaluated based on its final state \((X_f, \theta_f)\) and the path it took, represented as \((X_t, \theta_t)\) for \(t \in [0, T_f]\), where \(T_f\) is the final time step. Due to the shorter distances of the tasks we present, a maximum of 50 steps is allowed for the agent to navigate to its final destination, upon which the simulation episode is automatically stopped.

\subsection{Freeze-Time}
As the dynamic obstacles' positions are update on every simulation tick, differences in hardware performance - and hence inference speed - can lead to big differences in simulation results. To ensure that experiments are hardware-agnostic, we introduce the idea of "Freeze-Time" when conducting VLN experiments, where we pause the simulation when an agent is predicting the next action to take. This is a toggleable feature in our AdaSimulator, and can be switched off if future works wish to take inference speed into account when evaluating navigation performance.

\section{Method}

\subsection{AdaSimulator}

AdaSimulator is implemented as a standalone extension to IsaacSim, leveraging its physics engine and RTX Renderer. The simulator automatically sets up all necessary environment components when loading a scene:

\begin{itemize}
    \item Sets up collider meshes for the static obstacles
    \item Spawns a Jetbot at \((X_0, \theta_0)\)
    \item Sets up camera render products for generating observations
    \item Loads environment lighting rigs
    \item Loads humans at  \((X'_0, \theta'_0)\) and sets up their animation graphs
\end{itemize}

All simulation scenarios use a two-wheeled NVIDIA Jetbot, controlled via differential controllers for physics-based movement. All egocentric observations are rendered through IsaacSim's Replicator Core, using the Jetbot's attached camera for render perspective. Dynamic human obstacles are introduced into the environment via a customized version of the \texttt{omni.anim.people} \cite{nvidia2024omnianimpeople} extension. A ROS2 interface is also provided, allowing agents to extract RGB-D observations from the simulator and send control commands.

The simulator can be run in GUI mode for full visibility of the navigation episodes and manual input of robot commands, or in headless mode for optimal training speed.

\subsection{AdaR2R (Sample)}

\begin{figure}[t]
    \centering
    \includegraphics[height=1.4\linewidth]{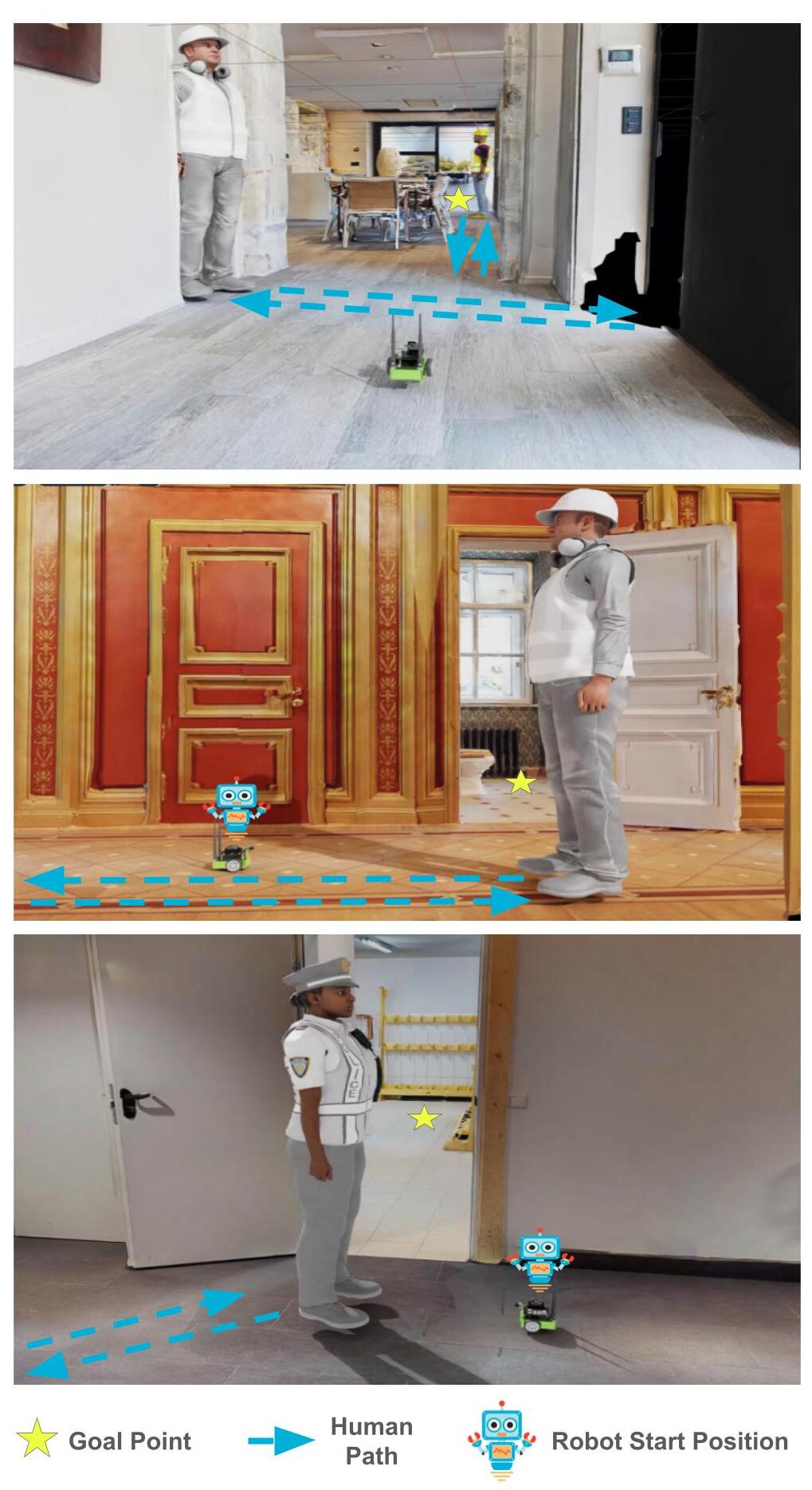}
    \caption{Top: Environment the 9 navigation episodes were conducted in. Humans loop along the indicated paths infinitely throughtout a navigation episode. Note that the paths have been deliberately chosen to interfere with the optimal path the robot would take. }
    \label{fig:examples}
\end{figure}

AdaR2R (Sample) is an example dataset containing 9 navigation episodes across 3 HM3Dv2 example scenes \cite{Ramakrishnan_Gokaslan_Wijmans_Maksymets_Clegg_Turner_Undersander_Galuba_Westbury_Chang_2021}. Snapshots of these episode's environments and human obstacles are shown in Figure \ref{fig:examples}. It modifies the original R2R dataset format to include configurations for human spawn points, path waypoints, and movement parameters. The example configurations have been manually set up to include 1-2 humans per episode, and their paths waypoints are chosen such that they will directly interfere with straight-line paths between critical nodes provided in the reference path. However, these interferences are never permanent, and there will either always be an alternative route that curves around the obstacle, or the obstacle will eventually move away as part of its patrol.

The tasks are purposely made to be simple as the focus is on the human obstacles, with an average geodesic distance for each navigation episode is 5.84 meters.

As a sample, it serves as a reference for future works to establish new task variants of existing room-to-room datasets. The environment and robot both use triangular collider meshes with default offset values determined by IsaacSim.

\section{Experiments}

\subsection{Evaluation Protocol}

To give a baseline demonstration of the task and use of the simulator, we evaluate a baseline agent's ability to navigate to its goal and avoid collision with both humans and environmental obstacles. Established evaluation metrics for VLN tasks typically focus on the navigation performance of agents \cite{Anderson_Chang_Chaplot_Dosovitskiy_Gupta_Koltun_Kosecka_Malik_Mottaghi_Savva_et_al_2018, Anderson_Wu_Teney_Bruce_Johnson_Sünderhauf_Reid_Gould_Hengel_2018, Yue_Zhou_Xie_Zhang_Yan_Yin_2024}. Due to our focus on introducing a new simulation framework rather then an agent, we will instead look at our baseline agent's collisions with environmental and human obstacles instead. Navigation Collisions (NC) records the ratio of the total amount of time an agent is in collision with either a human or static environmental obstacles (walls, furnitures etc.) to the total navigation time. We also break it down into Human Navigation Collisions (HNC) and Environmental Navigation Collisions. We will also do a qualitative analysis of our baseline agent's observations and actions for several navigation episodes.

Agents are limited to a maximum of 50 navigation steps per episode, owing to the shorter geodesic distances of our task compared to R2R and R2R-CE.

\subsection{Physics Setup}

The NVIDIA Jetbots were setup with differential controllers configured for wheel radius of 0.035 m and wheel base distance of 0.1 m.

\begin{figure}
    \centering
    \includegraphics[width=\columnwidth]{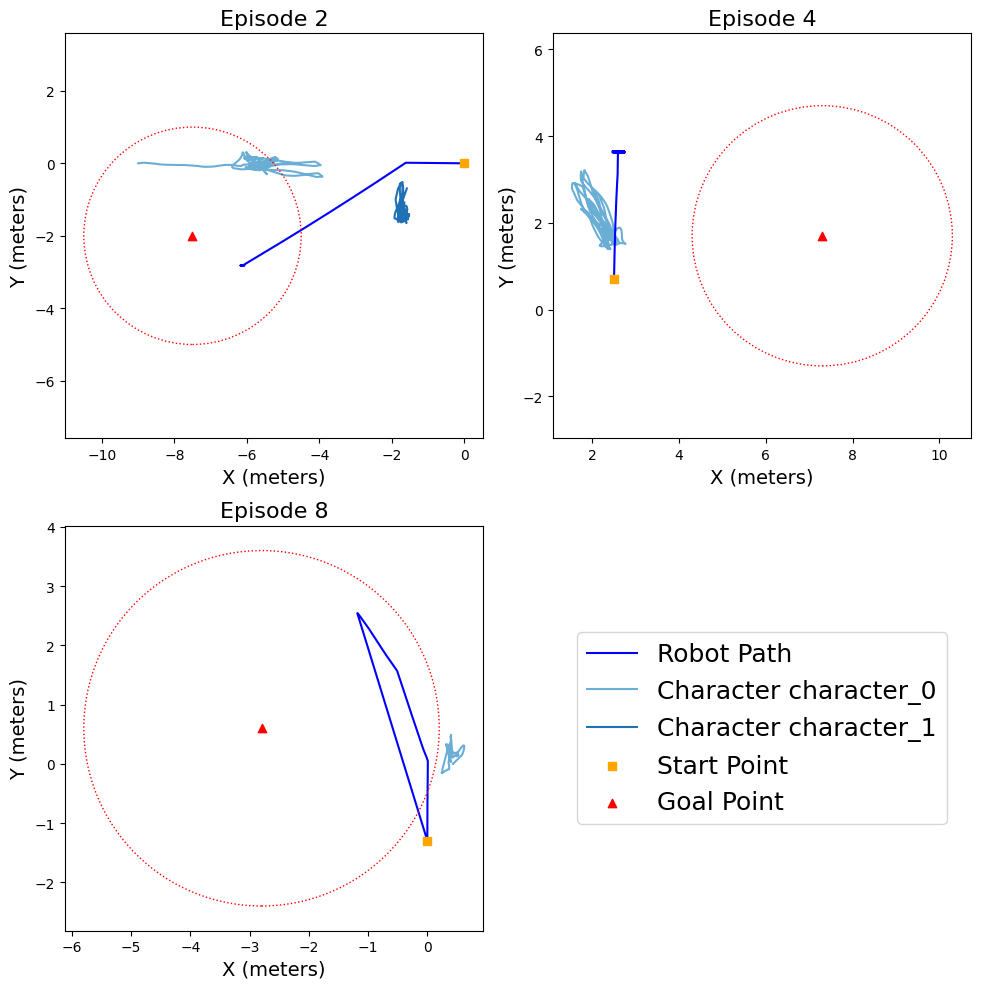}
    \caption{Top: Sample of paths (represented by lines) taken by robots and humans during simulation. Coordinate origins are based on X-Y provided in MP3D GLB files which have been scaled to 1 unit : 1 meter. In cases where the robot's line moves back-and-forth around a point, the robot has gotten stuck in collision with a wall.}
    \label{fig:output_plot}
\end{figure}

\subsection{Baseline GPT Agent}

We test a simple agent that uses the GPT-4o-mini \cite{openai2024gpt4omini} multi-modal foundational model. At each navigation step, the robot was told to generate semantic observations from the images, before generating a long-term plan to progress in the task. The robot then looks back on previous steps, then reasons about the logical next step required to continue along the plan, before finally making a decision. Predictions were limited to 200 output tokens per navigation step, and only RGB images used as input (after down-scaling from the render product's 1280x720 to 640x360 pixels) to reduce complexity and token usage. Implementation details are available in the project repository.

Tests are conducted zero-shot with no prior training of any sort on our dataset.

As seen in Figure \ref{fig:output_plot} and Table \ref{tab:nc_table}, collision rates in general are high, due to poor environmental parsing capabilities of our agent. In particular, we note that our agent frequently makes hallucinating observations which include:
\begin{itemize}
    \item Stating that paths ahead are clear even if they are facing a wall
    \item Stating that there are no humans or obstacles in front of them even if there are
    \item Hallucinating the instruction's objects in front of them
\end{itemize}

We note also that due to the full physics simulation of both the robot and the environment, it is much more difficult for robots to recover from colliding with walls compared to the HabitatSim simulators. Robots do not simply slide along the walls upon collision; rather, due to the nature of the robot's shape, it is common for the robot to roll over backwards as it attempts to move forward into a wall. Even if a robot does not flip, it is unable to turn effectively and hence is unable to escape as a "reverse" action is not defined. This makes it nearly impossible for a robot to get out of a static collision situation once it gets into one, which presents a new difficulty and layer of realism for such simulations. This is in contrast to other simulator like HabitatSim, which got around this issue by allowing robots to "slide" along the wall.

Although human collisions constitute a small proportion of the total collisions, this is primarily because humans continue on their paths and exit the collision zone after contact. As shown in \ref{fig:output_plot}, the agent makes little effort to navigate around human obstacles. We hypothesize that this behavior is due to the lack of realism in the human 3D models, causing the foundational model to fail to recognize them as obstacles.

\section{Conclusion and Future Work}

We presented AdaVLN, which extends the VLN-CE problem towards agent/robot navigation in dynamic environments featuring moving humans as dynamic obstacles. Alongside this, we introduced AdaSimulator, an extension of IsaacSim that facilitates the setup of fully physics-enabled simulations with realistic robots and animated 3D humans. Our baseline experiments demonstrate that the added complexity of our simulator enables more realistic evaluations and highlights the potential challenges of the new task. We aim to expand on this work by refining the simulation environment, generalizing the task formalization to broader dynamic environments, and developing agents capable of effectively navigating these complex scenarios.

This project is supported via the NVIDIA Academic Grant, including the A100 80GB GPU hardwares, NVIDIA IsaacSim software and Saturn Enterprise cloud.

\begin{table}[tb]
  \label{tab:nc_table}
  \scriptsize%
  \centering%
  \begin{tabu}{%
    r%
    *{3}{c}%
  }
  \toprule
   Episode & Environmental NC & Human NC & Combined NC \\
  \midrule
   1       & 0.77 & 0.01 & 0.78 \\
   2       & 0.69 & 0.01 & 0.70 \\
   3       & 0.91 & 0.01 & 0.91 \\
   4       & 0.93 & 0.00 & 0.93 \\
   5       & 0.86 & 0.00 & 0.86 \\
   6       & 0.76 & 0.00 & 0.76 \\
   7       & 0.00 & 0.00 & 0.00 \\
   8       & 0.71 & 0.08 & 0.78 \\
   9       & 0.00 & 0.00 & 0.00 \\
   \textbf{Average} & \textbf{0.63} & \textbf{0.01} & \textbf{0.64} \\
  \bottomrule
  \end{tabu}%

  \hspace{1cm}
  \caption{Normalized Collision (NC) values across different episodes. The combined navigation collision measures the total amount of timesteps a robot spends in a navigation episode while in collision with any object in the scene. Environmental and Human NC only considers collisions with static obstacles (like furnitures/wall) or humans respectively.}
\end{table}

\section{Acknowledgements}
This project is supported via the NVIDIA Academic Grant, including the A100 80GB GPU hardwares, NVIDIA IsaacSim software and Saturn Enterprise cloud.

\bibliographystyle{abbrv-doi}

\bibliography{main}

\end{document}